# PatentMiner: Patent Vacancy Mining via Context-enhanced and Knowledge-guided Graph Attention

Gaochen Wu[1], Bin Xu[1], Yuxin Qin[2], Fei Kong[3], Bangchang Liu[3], Hongwen Zhao[3], Dejie Chang[3]

[1,2]Computer Science and Technology, Tsinghua University, Beijing, China
[3]Beijing MoreHealth Technology Group Co. Ltd
[1]{wgc2019, xubin}@tsinghua.edu.cn, [2]{tyx16}@mail.tsinghua.edu.cn
[3]{kongfei, liubangchang, zhaohongwen, changdejie}@miao.cn

**Abstract.** Although there are a small number of work to conduct patent research by building knowledge graph, but without constructing patent knowledge graph using patent documents and combining latest natural language processing methods to mine hidden rich semantic relationships in existing patents and predict new possible patents. In this paper, we propose a new patent vacancy prediction approach named **PatentMiner** to mine rich semantic knowledge and predict new potential patents based on knowledge graph (**KG**) and graph attention mechanism. Firstly, patent knowledge graph over time (e.g. year) is constructed by carrying out named entity recognition and relation extraction from patent documents. Secondly, Common Neighbor Method (**CNM**), Graph Attention Networks (**GAT**) and Context-enhanced Graph Attention Networks (**CGAT**) are proposed to perform link prediction in the constructed knowledge graph to dig out the potential triples. Finally, patents are defined on the knowledge graph by means of co-occurrence relationship, that is, each patent is represented as a fully connected subgraph containing all its entities and co-occurrence relationships of the patent in the knowledge graph; Furthermore, we propose a new patent prediction task which predicts a fully connected subgraph with newly added prediction links as a new patent. The experimental results demonstrate that our proposed patent prediction approach can correctly predict new patents and Context-enhanced Graph Attention Networks is much better than the baseline. Meanwhile, our proposed patent vacancy prediction task still has significant room to improve.

**Keywords:** knowledge graph, Graph Attention Networks, link prediction, co-occurrence relationship.

## 1 Introduction

Patent is a kind of intellectual property, which endows inventors with the exclusive right to the invention within a certain period, so that it can be widely used to promote progress of science and technology and development of industry. Patents are valuable knowledge and technical information resources with characteristics of large quantity and wide content. Therefore, research on patents has very important theoretical value and high-practical significance to scientific research and enterprise development (Jokanovic et al., 2017).

Patent documents contain a wealth of entities and relationships, which enable us to construct the patent knowledge graph with knowledge extraction (Schlichtkrull et al., 2018), so as to dig out hidden semantic relationships in the existing patents by virtue of knowledge graph and graph neural network (Veličković et al., 2019; Zhou et al., 2018). Sarica et al. (2019) attempted to use natural language processing techniques to build an engineered knowledge graph with patent database as data resources and thereby provide convenience for patent retrieval.

However, they did not explore and utilize new latest technologies of knowledge graph, for example, TransE (Bordes et al., 2013) and graph neural networks (Velickovic et al., 2018), to deeply mine potential rich semantic relationships hidden in patents. Xu et al. (2015) used Freebase (Bollacker et al., 2008) and Mesh word table to create knowledge graph in the field of lung cancer, and then tagged patent literatures with the knowledge graph, and made emerging technology prediction based on networks between labels. However, they did not directly construct a patent knowledge graph based on patent documents to predict emerging technologies, nor did take advantage of co-occurrence relationship (Surwase et al., 2011) to define patents in the knowledge graph. To sum up, the research direction of new technology prediction based on patent knowledge graph is almost blank.

It is of great significance and value to determine directions of scientific research and development strategies of enterprises to predict potential technical blank points that have not been applied for patents by mining existing massive patent documents. In this paper, we propose a patent vacancy prediction method called **PatentMiner** based on knowledge graph and graph neural networks to explore rich semantic information hidden in patents and predict potential possible new patents. Firstly, based on patent documents, the patent knowledge graph that changes with year is constructed through named entity recognition and relation extraction. Then, Common Neighbor Method (Taskar et al., 2003), Graph Attention Networks (Velickovic et al., 2018; Vaswani et al., 2017) and Context-enhanced (Peters et al., 2018; Devlin et al., 2019; Raford et al., 2018) Graph Attention Networks are proposed for performing link prediction on the constructed knowledge graph, so as to dig out potential triples which currently exist but have not been found. Finally, we define a patent in patent knowledge graph by using co-occurrence relationship, that is, each patent is represented as a fully-connected subgraph containing all its entities and co-occurrence relationships in the patent knowledge graph. In addition, we propose a patent vacancy prediction task which predicts a fully-connected sub-graph with newly added prediction links as a new potential possible patent.

In order to demonstrate the effectiveness of our proposed patent prediction approach in this work, we take advantage of the USPTO patent database and select patent documents from 2010 to 2019 in the field of electronic communication for carrying out a series of experiments. Experimental results show our proposed PatentMiner can accurately predict new patents, especially, prediction accuracy of Context-enhanced Graph Attention Networks (CGAT) is superior to the baseline (e.g. Common Neighbor Method), detailed in Table 3 and section 4.4, which fully demonstrate the capability of our proposed approach predicting new potential patents.

In summary, this paper has made the following contributions: **(1)** We propose a patent prediction approach called PatentMiner by combining knowledge graph and graph attention networks, which can accurately predict new potential possible patents. **(2)** We define a patent in patent knowledge graph by using co-occurrence relationship, that is, each patent is represented as a fully-connected subgraph of the knowledge graph containing all its entities and co-occurrence relationships, which quantifies accurately patent in knowledge graph. **(3)** A patent vacancy prediction task is proposed on patent knowledge graph, which predicts a fully connected subgraph containing newly added prediction links as a new possible patent. Meanwhile, we provide a baseline called common neighbor method and develop a state-of-the-art model called context-enhanced graph attention networks. **(4)** Experimental results demonstrate that our proposed approach (PatentMiner) can accurately predict new potential patents. The more important observation is that our proposed context-enhanced graph attention networks is much better than the baseline. Moreover, there is still significant room for improvement over the patent prediction task.

## 2   Related Work

Patent clustering and automatic classification are common topics in patent research. S. Jun (2014) constructed a combinatorial classification system by using data reduction and K-means clustering to solve the problem of sparseness in document clustering. Wu et al. (2016) used self-organizing mapping and support vector machine classification models to design a patent classification approach based on patent quality.

Statistical and probabilistic models are widely used in patent analysis. R. Klinger et al. (2008) proposed an approach to extract names of organic compounds from scientific texts and patents by using conditional random field models and Bootstrapping method. M. Klallinger et al. (2015) developed a set of chemical substance recognition systems by combining conditional random field model and support vector machine, and using natural language processing methods such as rule matching and dictionary query. A. Suominen et al. (2017) analyzed 160,000 patents using implicit Dirichlet distribution (LDA) method, and classified them into different groups according to patent theme. They established knowledge portrait of company. Lee C et al. (2016) divided technology life cycle into several stages and used hidden Markov chain model to predict the probability that the technology included in a patent is in a specific stage in the life cycle.

In recent years, neural network methods have been widely used in patent research. Lee et al. (2017) used the number of patent application and combined deep belief neural networks to predict future performance of companies. A. Trappey et al. (2012) combined principal component analysis method and back propagation neural networks to improve analysis effect of patent quality and shorten time needed to determine and evaluate the quality of new patents. Paz-Marin et al. (2012) used evolutionary S-type unit neural networks and evolutionary product unit neural networks to predict research and development performance of European countries. B. Jokanovic et al. (2017) assessed economic value of patents based on different scientific and technical factors using over-limit machine.

Patent documents contain a large number of entities and various relationships between entities, which make patent documents good materials for constructing knowledge graph. Therefore, it will be a very attractive research direction to make use of knowledge graph to explore hidden rich semantic relationships in patent data. Saprica et al. (2019) attempted to use natural language processing methods to build an engineering knowledge graph using patent database as the data source, so as to provide convenience for patent retrieval. Xu et al., (2012) used Freebase (Bollacker et al., 2008) and Mesh word table to create knowledge graph in the field of lung cancer, and then tagged patent literatures with this graph, and finally made emerging technology prediction based on the networks between labels.

Although there are a small amount of research by building knowledge graph for studying patents, but we could not find work by combining patent knowledge graph and latest natural language processing methods, such as pretraining models (Peters., 2018; Devlin et al., 2019; Radfordet al., 2018), attention mechanism (Vaswani et al., 2017) and graph neural network (Velickovic et al., 2018), to mine rich semantic relationships and knowledge hidden in patents and then forecast new patents. In this paper, we propose a patent vacancy prediction approach called **PatentMiner** based on knowledge graph and graph attention networks to perform link prediction and forecast patents.

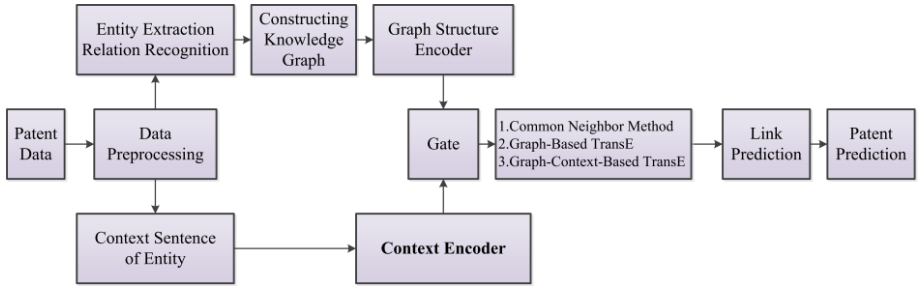

**Fig. 1.** PatentMiner - Our proposed patent vacancy prediction approach via context-enhanced and knowledge-guided graph attention mechanism by using gate mechanism to combine graph encoder and context encoder.

## 3    PatentMiner

The overall structure of our proposed patent prediction approach named PatentMiner is shown in Figure 1. Our approach consists of the following four steps: (1) Downloading and cleaning patent data, then constructing the patent knowledge graph through named entity recognition (NER) and relation extraction. (2) Embedding representations of entities and relationships using graph structure encoder (e.g.,GAT) and contextual text encoder(e.g., BERT). Studying link prediction approaches to mine potential triples in the graph. (4) Representing a patent as a fully-connected subgraph containing all its entities and co-occurrence relationships in the knowledge graph by using the co-occurrence relationship between entities, and forecasting a fully connected subgraph containing newly added prediction links as a new potential possible patent which may be applied in the future.

### 3.1 Patent Knowledge Graph Construction

Patent documents contain both structured and unstructured information, and research on patents includes analysis on these two parts. In this study, we firstly collect USPTO patent data in a certain period of time (e.g., from 2010 to 2019), and divide these patents according to the year of patent application into groups. Then entities and relationships in these patents are extracted and added to the knowledge graph by year. Finally, we build a patent knowledge graph KG(t) of different years for the following link and patent prediction tasks.

**Entity Recognition.** Named entity recognition refers to identification of entities with specific meaning in text, such as name of person, name of organization, name of location, etc. In this paper, domain entities are extracted from patent documents in the way of domain dictionary comparison.

**Relation Extraction.** Relation extraction is the extraction of semantic relations between two or more entities from text. In this work, based on co-occurrence relations, each patent is represented as a fully-connected sub-graph containing all its entities and co-occurrence relations in the patent knowledge graph, and we forecast a new possible patent as a fully-connected subgraph with newly added predictive co-occurrence links. Therefore, we firstly study the most important co-occurrence relationship between entities, that is, if two entities appear in a same patent document, we consider these two entities having a co-occurrence relationship. This method borrows the concept of co-citation proposed by Small Henry et al., (2013), an American intelligence scientist, that is to say, when two literatures are cited by the same paper, there is an internal relationship between them (Small Henry et al., 1977; Small Henry, Boyack Kevin et al., 2013; Small Henry, Boyack Kevin et al., 2014).

### 3.2 Graph and Context Embedding

After the construction of the patent knowledge graph in section 3.1, in order to carry out link prediction on the graph to dig out potential triples, and then to further conduct patent prediction, it is necessary to encode the entities and relations in the graph. Contextual information and graph structure are crucial to accurately represent entities. In this study, graph structure encoder and context encoder are used to capture information of entities in the graph structure and the patent context text, respectively.

**Graph Encoder.** The graph structure encoder assigns a vector representation to each entity based on its position in the graph and the characteristic representation of its adjacent entities. In this study, we use graph attention mechanism to feature weighted sum of adjacent nodes as the vector representation of the target node. The details are shown as follows.

The constructed patent knowledge graph is formulated with a list of tuples $\left(e_i^{head}, r_i, e_i^{tail}\right)$ composing of a head entity $e_i^{head}$, a tail entity $e_i^{tail}$, and their relation $r_i$. We randomly initialize vectors $\vec{e_i}$ and $\vec{r_i}$ for $e_i$ and $r_i$ respectively.

Considering a single layer of graph attention on the graph, assuming that the input of this layer is a set of entity node $h = \{\vec{h_1}, \vec{h_2}, ..., \vec{h_N}\}, \vec{h_i} \in R^F$, the output is a set of new features in Eqn. (4.b) $\bar{h} = \{\vec{h_1'}, \vec{h_2'}, ..., \vec{h_N'}\}, \vec{h_i'} \in R^{F'}$, the input and output for here may have different dimensions $F$ and $F'$. In order to convert input features $h$ into output features $\bar{h}$, a general linear transformation is required, which is represented by a weight matrix $W_s \in R^{F' \times F}$. The importance score of node j to node i can be calculated using a unified attention mechanism as in Eqn. (1) and Eqn. (3):

$$s_{ij} = a\left(W_s \vec{h_i}, W_s \vec{h_j}\right) \quad (1)$$

In this paper, $a$ in Eqn. (1) is a single-layer network and its parameters are expressed in terms of $\vec{a} \in R^{2F'}$. Use function softmax to normalize the importance score $s_{ij}$ in Eqn. (1) to obtain the attention weight $\alpha_{ij}$ as in Eqn. (2).

$$\alpha_{ij} = \text{softmax}(s_{ij}) = \frac{\exp(s_{ij})}{\sum_{k \in N_i} \exp(s_{ik})} \quad (2)$$

Where $s_{ij}$ is the result of a single-layer network with $\vec{a}$ as the parameters and LeakyReLU as the activation function over the concatenation of $\vec{h_i} \oplus \vec{h_j}$ of $\vec{h_i}$ and $\vec{h_j}$ as in Eqn. (3):

$$s_{ij} = \text{LeakyReLU}\left(\vec{a}^T \left[W_s \vec{h_i} \oplus W_s \vec{h_j}\right]\right) \quad (3)$$

Where $W_s$ is a learnable weight matrix.

Finally, the representation of the node $i$ in the patent knowledge graph can be calculated as in Eqn. (4):

$$\begin{aligned} \vec{h_i'} &= Sigmoid\left\{\sum_{j \in N_i} \alpha_{ij}\left(W_s \vec{h_j}\right)\right\} \quad (a) \\ \bar{h} &= \{\vec{h_1'}, \vec{h_2'}, ..., \vec{h_N'}\}, \vec{h_i'} \in R^{F'} \quad (b) \end{aligned} \quad (4)$$

**Context Encoder.** Each entity $e_i$ corresponding to the node $i$ in the constructed patent knowledge graph in section 3.1 appears in a certain context text. It is assumed that the sentence containing the entity $e_i$ is represented as $[w_1, w_2, ..., w_l]$, where $l$ is the length of the sentence and $w_i$ is the $i\text{-}th$ word in the sentence. In this study, BERT (Devlin et al., 2019) is used to encode the sentence to obtain the contextual representation of the entity $e$, and the vectors of each word in the sentence are represented as $[H_1, H_2, ..., H_l] \in R^d$, where $d$ is the dimension of the vector and $H_i$ represents the hidden state of $w_i$. Then we compute a bilinear attention weight for each word $w_i$ as in Eqn. (5).

$$\mu_i = e^T W_f H_i, \mu' = \text{Soft}\max(\mu) \tag{5}$$

Where $W_f$ is a bilinear matrix. We finally get the context representation as in Eqn. (6):

$$\bar{e} = \mu'^T H_i \tag{6}$$

**Gate Mechanism.** We use gate mechanism as in (Wang et al., 2019) to combine the graph-encoded representation $\bar{h}$ calculated in Eqn. (4.b) with the context-encoded representation $\bar{e}$ obtained in Eqn. (6) to form the final representation $E$ of the entity $e$ as in Eqn. (7):

$$\begin{aligned} g_e &= Sigmoid(\overleftrightarrow{g}_e), \\ E &= g_e \odot \bar{h} + (1 - g_e) \odot \bar{e} \end{aligned} \tag{7}$$

Where $g_e$ is an entity-dependent gate function of which each element is in [0,1], $\overleftrightarrow{g}_e$ is a learnable parameter for each entity $e$, $\odot$ is an element-wise multiplication.

### 3.3 Link Prediction

In order to predict new potential possible patents on the constructed patent knowledge graph, we first make link prediction on the graph to dig out potential triples.

**Link Prediction Task.** In this work, the link prediction task on the knowledge graph is defined in the following way: given the patent document data up to time T (e.g., 2015), the patent knowledge graph at time T is constructed and contains entities and relations extracted from existing patents. The task of link prediction is to infer missing links (links that exist but are not observed currently) from existing nodes and edges, which represent possible emerging technologies in the future. It is feasible to make link prediction on the patent knowledge graph, because the entities and relations contained in the graph will increase gradually with the passage of time, and the existing entities and relations will not disappear.

Collecting patent data in a certain period of time (e.g., from 2010 to 2019), dividing these patents according to the year of patent application, extracting knowledge from these patent documents and adding them to the knowledge graph according to the year, finally the patent knowledge graph changing with year can be obtained.

**Common Neighbor Method.** The common neighbor method is based on assumption that if two people in a social network have many public acquaintances, they are more likely to meet each other than two people without any public contact. Therefore, the more neighbors two nodes in knowledge graph have in common, the higher the probability that there is a potential edge between them.

Considering two nodes $x$ and $y$ in the graph, and assuming that the sets of their neighbors are $\Gamma x$ and $\Gamma y$ respectively. A scoring function is defined as $s(x, y)$, representing the probability of a predicted link between $x$ and $y$. In the common neighbor method, $s(x, y)$ is the number of common neighbor nodes between $x$ and $y$, which is formally expressed as in Eqn. (8):

$$s(x, y) = |\Gamma x \cap \Gamma y| \tag{8}$$

A reasonable threshold can be set by analyzing the structure of the graph as $\zeta$, such as counting exit degree and entry degree of nodes. When $s(x, y) > \zeta$, it is assumed that there is a predictive link between $x$ and $y$.

**TransE.** The fundamental principle of TransE (Bordes Antoine et al., 2013) is to represent both entities and relations as vectors, and to transform the problem of judging the existence of a tuple to verify whether the arithmetic expressions between entities and relations are valid. For example, for a triplet $(h, r, t)$, the vectors $h$ and $t$ representing head entity and tail entity, and the vector $r$ representing the relation should satisfy $h + r \approx t$. The loss function of this model can be defined as in Eqn. (9):

$$L = \sum_{(h,r,t) \in K} \sum_{(h',r,t') \in K'} \max\left(0, \gamma + d(h+r,t) - d(h'+r,t')\right) \tag{9}$$

Where $\gamma$ is a positive constant and $d(u, v) = |u - v|_2^2$ is a distance function. After the model is optimized according to this loss function, each pair of triples $(h, r, t)$ can be scored using the model. The higher the score obtains, the more likely this tuple $(h, r, t)$ is to be true.

### 3.4 Patent Prediction

On the basis of link prediction, we can also forecast the future possible new patents.

**Patent Definition.** According to the co-occurrence relationship between entities, for a patent $P$, assuming that it contains a set of entities $E = \{e_1, e_2, ..., e_n\}$, with a total of $n$ entities, therefore, the patent $P$ in the patent knowledge graph should be expressed as a fully connected subgraph containing $n$ nodes and $R = \left\{\frac{n(n-1)}{2}\right\}$ co-occurrence relations, so the patent $P$ can be represented as $P = G(E, R)$, where $E$ is the entity set of the patent, $R$ is the set of all co-occurrence relationships on the entity set of the patent $P$.

**Patent Prediction Task.** For a patent knowledge graph of time $t$, link prediction is first performed on the graph, and then all predicted links are added to the graph to obtain a new patent knowledge graph. The new graph is likely to include some new patents that each is represented as a maximal fully connected subgraph with at least one predicted link. According to the patent definition, we can identify these predictive new patents and compare them with patents in subsequent years $t + \Delta t$ to confirm the accuracy of our proposed models forecasting new patents.

Table 1. Statistics of Constructed Dynamic accumulative Patent Knowledge Graph from 2010 to 2019 (year)

| Cut-off Year | Patents | Entities | Tuples |
|---|---|---|---|
| 2010 | 16990 | 2401 | 120132 |
| 2011 | 17787 | 3578 | 151209 |
| 2012 | 20642 | 4698 | 169823 |
| 2013 | 23503 | 5733 | 180217 |
| 2014 | 28716 | 6817 | 197821 |
| 2015 | 29473 | 7284 | 206412 |
| 2016 | 31464 | 7905 | 220719 |
| 2017 | 33697 | 8601 | 258121 |
| 2018 | 32981 | 9054 | 273930 |
| 2019 | 35742 | 9901 | 291203 |

## 4 Experiments and Results

We verify effectiveness of our proposed approach in this part. Firstly, we select and download the USPTO patents from 2010 to 2019 in the field of electronic communication, and preprocess the patent data. Secondly, based on the patent data, we construct the patent knowledge graph changing with year. Finally, link prediction and patent prediction are carried out on the constructed patent knowledge graph.

### 4.1 Patent Data Preprocessing

We firstly download patent data including patent summary and CPC classification number from PatentsView, a database of the U.S. Patent and Trademark Office, and align the downloaded table data according to the unique patent number. Then, we select the patent documents containing H04L in the classification number from 2010 to 2019 according to the year and classification number. Finally, the title and abstract parts of each patent document are extracted and combined to form a separate text file. The number of patents by year are shown in Table 1.

### 4.2 Constructing Patent KG(Year)

We download a glossary of terms for electronic communications from the U.S. Federal Standards website, which cover 19,803 domain entities in the field of electronic communications. Starting with the patent data in 2010, the patent documents are compared with glossary year by year to obtain knowledge graphs with an increasing number of entities and relationships. The detailed statistics of the knowledge graphs by different years are shown in Table 1.

Table 2. Results of Link Prediction

| Cut-off Year | Link Prediction (Numeber of New Links) | | | Accuracy of Link Prediction (%) | | |
|---|---|---|---|---|---|---|
| | CNM | GAT | CGAT | CNM | GAT | CGAT |
| 2010 | 9821 | 13021 | 16001 | 12.12 | 15.92 | 21.73 |
| 2011 | 10215 | 16723 | 18902 | 13.34 | 16.47 | 22.63 |
| 2012 | 10928 | 20812 | 21982 | 13.91 | 18.03 | 24.85 |
| 2013 | 11384 | 26914 | 27031 | 14.14 | 19.85 | 25.91 |
| 2014 | 11892 | 32312 | 31019 | 15.08 | 20.54 | 26.14 |
| 2015 | 12309 | 36912 | 38632 | 15.87 | 21.32 | 27.13 |
| 2016 | 13018 | 42453 | 45812 | 16.82 | 22.71 | 28.95 |
| 2017 | 14219 | 48912 | 51238 | 17.02 | 24.56 | 29.73 |
| 2018 | 16921 | 56834 | 64123 | 17.82 | 25.64 | 32.41 |
| 2019 | 18388 | 62981 | 72943 | 18.54 | 26.01 | 33.81 |

### 4.3 Link Prediction Results

In this study, we use link prediction accuracy and patent prediction accuracy to measure the performance of our proposed methods. For the predicted new links, we examine whether they accurately appear in the graphs constructed using patents which are applied in subsequent years. For example, considering the constructed graph of 2010 (e.g., KG(2010)), which means that the graph is constructed using all patents up to 2010, supposing the set of new links predicted by the proposed approach over the graph of 2010 are represented as $R(2010)$ and the whole link set of the graph of the year (e.g., KG(Year)) as $R(Year)$, then the prediction link accuracy is calculated as in Eqn. (10):

$$a_{link}(2010 \mid Year) = \frac{\mid R(Year) \cap R(2010) \mid}{\mid R(2010) \mid} \quad (10)$$

Where $\mid R(Year) \cap R(2010) \mid$ represents the number of accurate prediction links.

In order to investigate effectiveness of our proposed approach, we use common neighbor method (**CNM**) as the baseline and verify graph attention networks (**GAT**) and context-enhanced graph attention networks (**CGAT**).

**CNM.** We count the maximum number of common neighbors of each pair of nodes in the graph as $M$. If a pair of nodes does not have a link connection, but their common neighbors $m \geq \left\lceil \dfrac{M}{2} \right\rceil$, a new link is considered to exist between this pair of nodes.

**GAT.** Firstly, we utilize a graph structure encoder with the attention mechanism to encode the entities and relations in the graph, and then the TransE method is used to train the prediction model. Finally, we take advantage of the model to score and forecast the potential links.

**CGAT.** On the basis of **GAT**, context encoder is added to enrich the representations of entities, such as BERT in this paper. We introduce a gate function to combine the node representations learned by graph encoder and context encoder to form a comprehensive representation for each entity.

Table 3. Results of Patent Prediction

| Cut-off Year | Patent Prediction (Numeber of New Patents) | | | Accuracy of Patent Prediction (%) | | |
|---|---|---|---|---|---|---|
| | CNM | GAT | CGAT | CNM | GAT | CGAT |
| 2010 | 2081 | 3083 | 3729 | 7.32 | 10.23 | 11.23 |
| 2011 | 2871 | 3612 | 4011 | 8.14 | 10.96 | 11.92 |
| 2012 | 3067 | 3902 | 5874 | 8.95 | 11.32 | 12.37 |
| 2013 | 3571 | 4205 | 6920 | 10.21 | 11.47 | 12.78 |
| 2014 | 4018 | 4923 | 7810 | 11.94 | 12.03 | 14.01 |
| 2015 | 5219 | 5612 | 8451 | 12.29 | 12.57 | 14.81 |
| 2016 | 5598 | 6472 | 9012 | 12.63 | 12.83 | 15.23 |
| 2017 | 6201 | 7201 | 10983 | 12.91 | 13.42 | 16.02 |
| 2018 | 7034 | 8312 | 12394 | 13.81 | 14.71 | 16.72 |
| 2019 | 7561 | 9056 | 14834 | 14.05 | 15.23 | 17.31 |

According to the experimental results in Table 2, as the scale of the graph increases, the number of new links predicted by the three models quickly go up. Furthermore, the accuracy of link prediction improves year by year. Compared to the baseline, we observe that the number of new links found using GAT and CGAT increase significantly in every year from 2010 to 2019. More importantly, the link prediction accuracy of CGAT is significantly improved over the baseline on the same dataset, with an average increase of 4 points.

### 4.4 Patent Prediction Results

According to the definition of co-occurrence relationship, all entities and their co-occurrence relationships contained in each patent are represented as a fully connected subgraph in the knowledge graph. In this paper, the fully connected subgraph containing new predicted links is regarded as a new patent forecasted by the model.

Supposing the set of domain entities included in a new predicted patent $p = G(E, R)$ as $E$, if there is at least one patent $p' = G(E', R')$ applied in the future years, where its set of domain entities satisfy $E' \supseteq E$, then the predicted new patent $p$ is considered to be valid. If the set of all new forecasted patents is $P$, and the valid set of new patents is $P_0$, then the accuracy of patent prediction is calculated as in Eqn. (11).

$$a_{patent} = \frac{|P_0|}{|P|} \qquad (11)$$

Experimental results are shown in Table 3. With the scale of the graph increasing, we observe that the number of new patents predicted by the three models increases, and the patent prediction accuracy also improves year by year.

Furthermore, for the patent prediction task, CGAT using both contextual representation and graph representation is better than GAT only based on graph encoder. Moreover, the maximum prediction accuracy of CGAT reaches 17.31% in 2019 with 3.26 points higher than the baseline and 2.06 points higher than GAT. The experimental results demonstrate that our proposed approach can predict new potential possible patent.

## 5 Conclusion

In this study, we propose a new patent vacancy prediction approach called PatentMiner via patent knowledge graph and graph attention. We define a patent on knowledge graph by using co-occurrence relationships, and a patent prediction task is proposed to predict the fully connected subgraph containing new predictive links as a new patent. Experimental results demonstrate that our proposed approach can correctly predict new patents. Meanwhile, there is still much room for improvement on the patent prediction task.

## Acknowledgements

We would like to thank all anonymous reviewers for their thorough reviewing and providing constructive comments to improve this paper. This work was supported by the Ministry of Science and Technology via grant 2017YFB1401903 and 2018YFB1005101. This work was also supported by Beijing MoreHealth Technology Group Co. Ltd.